\pdfoutput=1

\documentclass[11pt]{article}
\usepackage{times,latexsym}
\usepackage{url}
\usepackage[T1]{fontenc}
\usepackage{booktabs}
\usepackage{enumitem}
\usepackage[english]{babel}
\usepackage{multirow}
\usepackage{natbib}
\usepackage{bbm}
\usepackage{xcolor}
\usepackage{framed}
\usepackage{pgfplots}
\usepackage{tikz}
\usepackage{caption}
\usepackage{subcaption}
\usepackage[utf8]{inputenc}
\usepackage{tikz}
\usepackage{amssymb}
\usepackage{makecell}
\usepackage{amsmath}
\usepackage{tabularx}
\usepackage{cuted}
\usepackage{floatrow}
\usepackage{colortbl}
\newfloatcommand{capbtabbox}{table}[][\FBwidth]
\usepackage[textsize=scriptsize]{todonotes}

\newcommand\gd[1]{}
\newcommand\af[1]{}
\newcommand{\greg}[1]{}
\newcommand\ab[1]{}
\newcommand{\adithya}[1]{}

\definecolor{bblue}{HTML}{4F81BD}
\definecolor{rred}{HTML}{C0504D}
\definecolor{ggreen}{HTML}{9BBB59}
\definecolor{ppurple}{HTML}{9F4C7C}

\usepackage{ACL2023}

\usepackage{times}
\usepackage{latexsym}

\usepackage[T1]{fontenc}

\usepackage[utf8]{inputenc}

\usepackage{microtype}

\usepackage{inconsolata}
\pgfplotsset{compat=1.17}

\title{Prompted Opinion Summarization with GPT-3.5}

\author{
    Adithya Bhaskar$^1$\\IIT Bombay \And Alexander R. Fabbri$^2$\\Salesforce AI \\$^1$\texttt{adithyabhaskar@cse.iitb.ac.in}\\$^2$\texttt{afabbri@salesforce.com}\\$^3$\texttt{gdurrett@cs.utexas.edu} \And Greg Durrett$^3$\\UT Austin
}

\begin{document}
\maketitle
\begin{abstract}
Large language models have shown impressive performance across a wide variety of tasks, including text summarization. 
In this paper, we show that this strong performance extends to opinion summarization. 
We explore several pipeline methods for applying GPT-3.5 to summarize a large collection of user reviews in a prompted fashion. To handle arbitrarily large numbers of user reviews, we explore recursive summarization as well as methods for selecting salient content to summarize through supervised clustering or extraction. 
On two datasets, an aspect-oriented summarization dataset of hotel reviews (SPACE) and a generic summarization dataset of Amazon and Yelp reviews (FewSum), we show that GPT-3.5 models achieve very strong performance in human evaluation. 
%
We argue that standard evaluation metrics do not reflect this, and introduce three new metrics targeting faithfulness, factuality, and genericity to contrast these different methods. 
\end{abstract}
\section{Introduction}
\begin{figure*}[t]
\centering
\includegraphics[width=0.95\linewidth,trim=7mm 90mm 20mm 30mm]{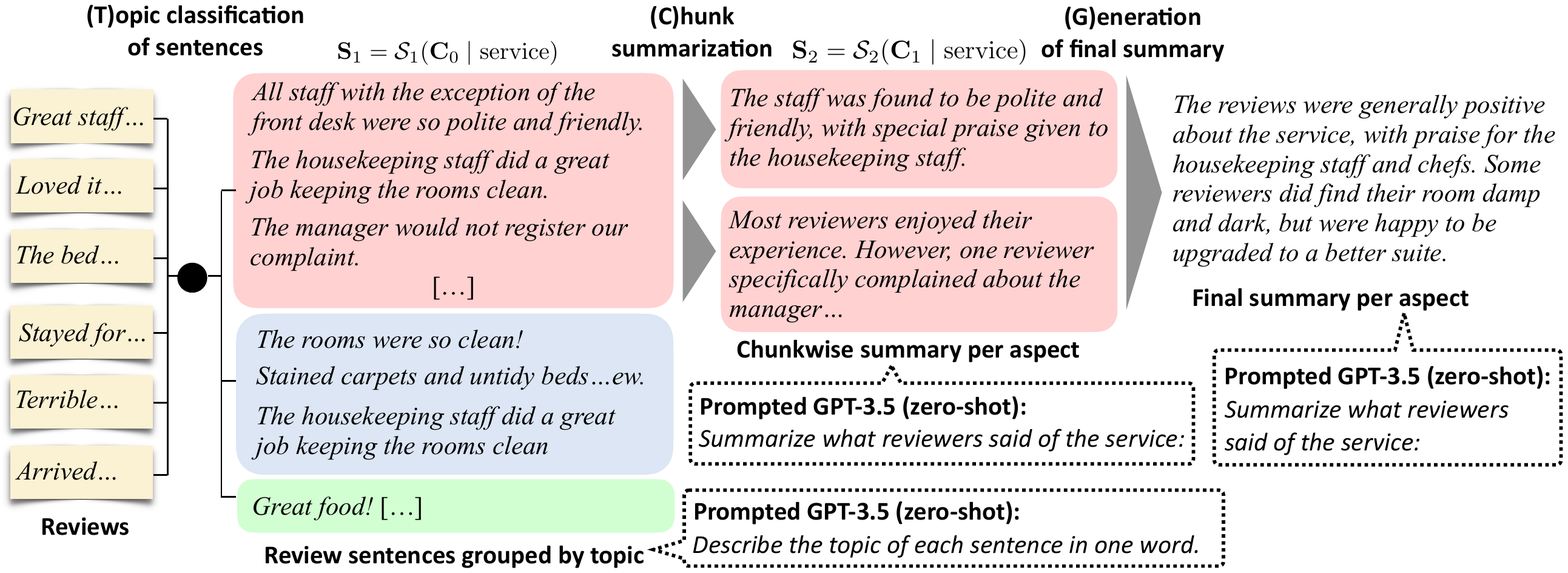}
\caption{Illustration of the TCG pipeline. Sentences are clustered based on the aspects closest to their topic (T step); examples are shown for \textcolor{blue}{rooms}, \textcolor{green}{food} and \textcolor{red}{service}. The relevant cluster is then repeatedly chunked and summarized until the combined length falls below 35 sentences (C step). A final round of GPT-3.5 summarization follows (G step).}
\label{fig:TCGI}
\end{figure*}
Recent years have seen several shifts in summarization research, from primarily extractive models \cite{LexRank, Ext1, Ext2, Ext3, Ext4} to abstractive models with copy mechanisms \cite{Copy1, Copy2, Copy4} to pre-trained models \cite{BERT, Abs1, BART, Pegasus, NG1}. GPT-3 \cite{GPT3,wu-books,saunders-critiques,goyal-gpt3} and GPT-4 represent another shift: they show excellent zero- and few-shot performance across a variety of text generation tasks. However, their capabilities have not been extensively benchmarked for opinion summarization. Unlike news, where extractive lead baselines are often highly effective, opinion summarization requires balancing contradictory opinions and a higher degree of abstraction to convey all of the viewpoints faithfully.

In this paper, we apply GPT-3.5, specifically the \texttt{text-davinci-002} model,\footnote{The most advanced model available at the time this work was being conducted.} to the task of opinion summarization, focusing on reviews of products, hotels, and businesses. 
Applying GPT-3.5 in this setting is not straightforward, as the combined length of the reviews or posts may exceed the model's maximum input length.
Furthermore, we find that certain styles of inputs can lead to GPT-3.5 simply echoing back an extract of the inputs.  
To mitigate these issues, we explore a family of pipelined approaches, specifically (1) filtering a subset of sentences with an extractive summarization model, (2) chunking with repeated summarization, and (3) review-score-based stratification. 
In the context of aspect-oriented summarization, we also explore the inclusion of a sentence-wise topic prediction and clustering step.

We show that our approaches yield high-quality summaries according to human evaluation.
The errors of the systems consist of subtle issues of balancing contradictory viewpoints and erroneous generalization of specific claims, which are not captured by metrics like ROUGE \cite{ROUGE} or BERTScore \cite{BERTScore}.
This result corroborates work calling for a re-examination of current metrics \cite{SummEval,tang2023understanding} and the need for fine-grained evaluation \cite{gehrmann2022repairing}.
%
We therefore introduce a set of metrics, using entailment as a proxy for support, to measure the \emph{factuality}, \emph{faithfulness}, and \emph{genericity} of produced summaries.
These metrics measure the extent of over-generalization of claims and misrepresentation of viewpoints while ensuring that summaries are not overly generic.

Our results show that basic prompted GPT-3.5 produces reasonably faithful and factual summaries when the input reviews are short (fewer than $1000$ words); more sophisticated techniques do not show much improvement. However, as the input size grows larger, repeated summarization leads GPT-3.5 to produce generalized and unfaithful selections of viewpoints relative to the first round.
We demonstrate that using QFSumm \cite{QFSumm}, an extractive summarization model, to filter out sentences prior to GPT-3.5 (instead of multi-level summarization) can slightly help with factuality and faithfulness. 
The resulting summaries also present a more specific selection of viewpoints but are generally shorter and use a higher proportion of common words. 
A topicwise clustering and filtering step pre-pended to the pipeline alleviates these issues while relinquishing a portion of the gains on factuality and faithfulness.\gd{I don't know about this punchline at the end of the intro. IMO these results aren't that sharp, particularly factuality on the human eval}\ab{Removed 'significantly'.}
\par
Our main contributions are: (1) We introduce two approaches to long-form opinion summarization with GPT-3.5, namely, hierarchical GPT-3.5 summarization with chunking, and pre-extraction with an extractive summarization model. (2) We establish the strength of these approaches with a human study and demonstrate the need for objective and automatic means of evaluation. (3) We develop three entailment-based metrics for factuality, faithfulness, and genericity that are better suited to evaluate extremely fluent summaries as compared to metrics based on $n$-gram matching. The relevant artifacts and code for this work are publicly available and can be found at \url{https://github.com/testzer0/ZS-Summ-GPT3/}.

\section{Motivation and Problem Setting}
Review summarization involves the summarization of the text of multiple reviews of a given product or service into a coherent synopsis. More formally, given a set of reviews $\mathcal R = \{R_i\}_{i=1}^n$ with the review $R_i$ consisting of $l_i$ sentences $\{r_{ij}\}_{j=1}^{l_i}$, we define a \textit{summarization system} $\mathcal S$ to be a function that takes as input the combined reviews $C$ and then produces $k$ output sentences $S = \{s_i\}_{i=1}^k$, written as $S = \mathcal S(C)$, where $C \equiv \texttt{combine}(\mathcal R)$ is typically obtained by concatenating the review sentences. We use the notation \texttt{combine} to refer to the combination of both sentences and reviews.

We can also instantiate this pipeline for \emph{aspect-oriented review summarization}, which involves the summarization of multiple reviews conditioned on an aspect $a$ (such as \emph{`cleanliness'}). In particular, the summarization is written as $S = \mathcal S(C \mid a)$.
We consider aspect-agnostic review summarization as a special case of aspect-oriented review summarization with the aspect \emph{`none'} for notational simplicity.

\subsection{Desiderata} 
\label{sub:DD}
Opinion summaries should demonstrate three key characteristics.

First, the summaries should also be \textbf{faithful}, i.e., select the most subjectively important viewpoints with the largest consensus. For instance, if five reviews raised the issue of small rooms while eight complained about dusty carpets, the choice (due to a limited output size) to discuss the latter over the former would be considered faithful. Thus, faithfulness is about careful management of the word budget given constrained output length.

The summaries should also be \textbf{factual}, i.e., report information grounded in statements that actually do appear in the set of reviews, without containing extrinsic hallucinations. For instance, if five reviews found hotel rooms to be small, but three found them large, the statement \emph{The rooms were large} is considered factual despite the viewpoint being in the minority. By contrast, \emph{A pipe burst and flooded my room} is unfactual if this is never actually reported in the reviews.

Finally, the summaries should be \textbf{relevant}: the points raised in them should only discuss topics relevant to the specified aspect. For example, in a summary about the cleanliness of a hotel room, bad food should be omitted even if it was frequently brought up in the reviews.
\subsection{Framework}  Based on the desiderata, we need to ensure that the summaries represent all of the reviews; however they are too many in number and too long in combined length. We, therefore, define a \textit{summarization pipeline} to be a series of summarization systems $\mathcal S_1, \cdots, \mathcal S_m$ where each system takes as input the condensed results of the previous system.
Specifically,
$$S_0 = \mathcal R,\ \ C_i = \texttt{combine} (S_{i-1}),\ \ S_i = \mathcal S_i(C_{i})$$ 

\begin{table}[t]
\centering
\includegraphics[width=0.95\linewidth]{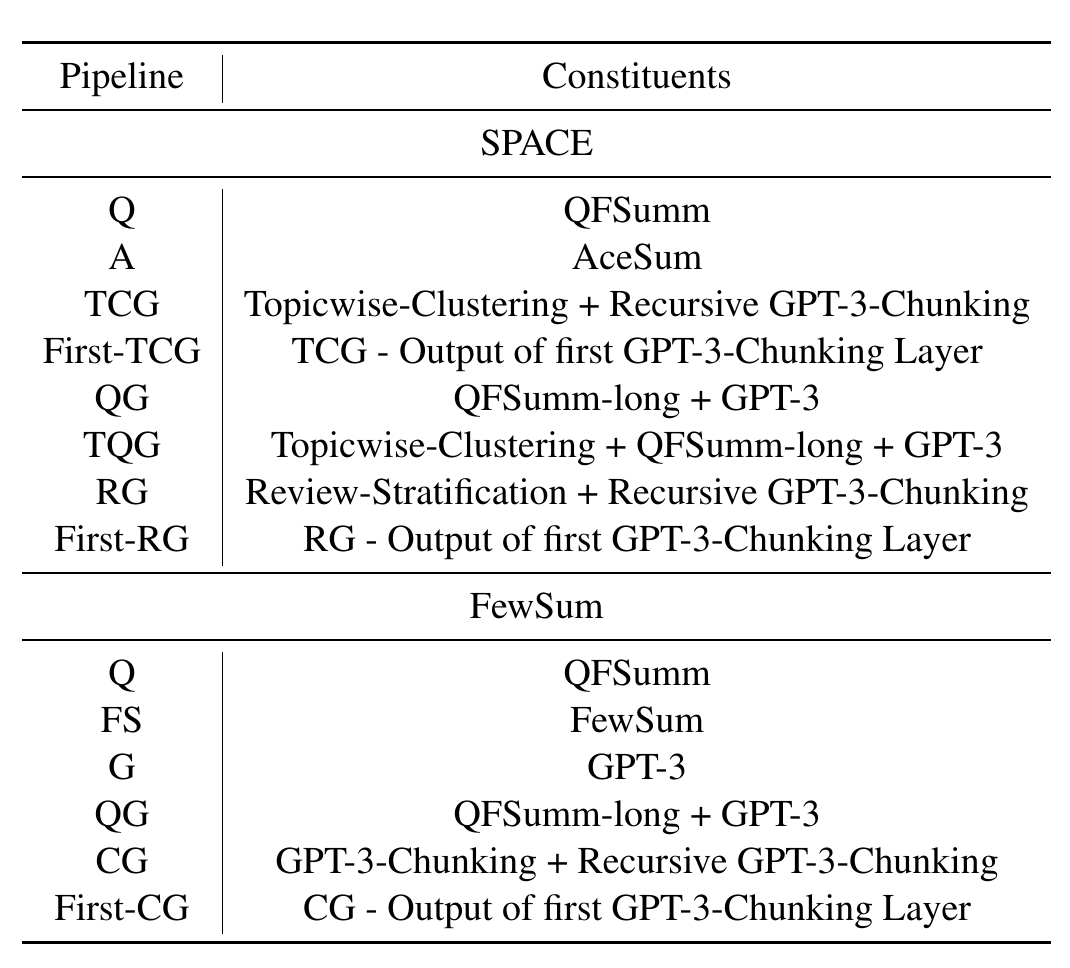}
\caption{The pipelines compared for SPACE and FewSum, and their constituents.}
\label{table:PP}
\end{table}

We showcase an example pipeline in Figure~\ref{fig:TCGI}, with one stage extracting the relevant sentences from the reviews and the next summarizing the extracted sentences.

\section{GPT-3.5 Summarization Pipelines}
The components of our summarization pipelines may be broadly categorized into \emph{extractors} and \emph{summarizers}, which we describe next. More details can be found in Appendix~\ref{ap:PD}. First, \textit{extractors} select relevant parts of a set of reviews, optionally conditioned on an aspect. Our extractors include:
\paragraph{GPT-3.5 Topic Clustering (T)} We prompt GPT-3.5 to produce a single word topic for each sentence, which we map to the closest aspect with GloVe \cite{GLOVE} similarity. This defines a set of sentences to be used for aspect-based summarization.  This step is only used for pipelines on SPACE, as FewSum is aspect-agnostic.
\paragraph{QFSumm-long (Q)} We use the aspect-specific \textit{extractive} summarization model introduced in \cite{QFSumm} to extract up to $35$ most relevant sentences from the input text. QFSumm was designed to allow extremely long inputs, and thus no truncation is required at this stage.

\begin{figure}[t]
\includegraphics[width=0.95\linewidth]{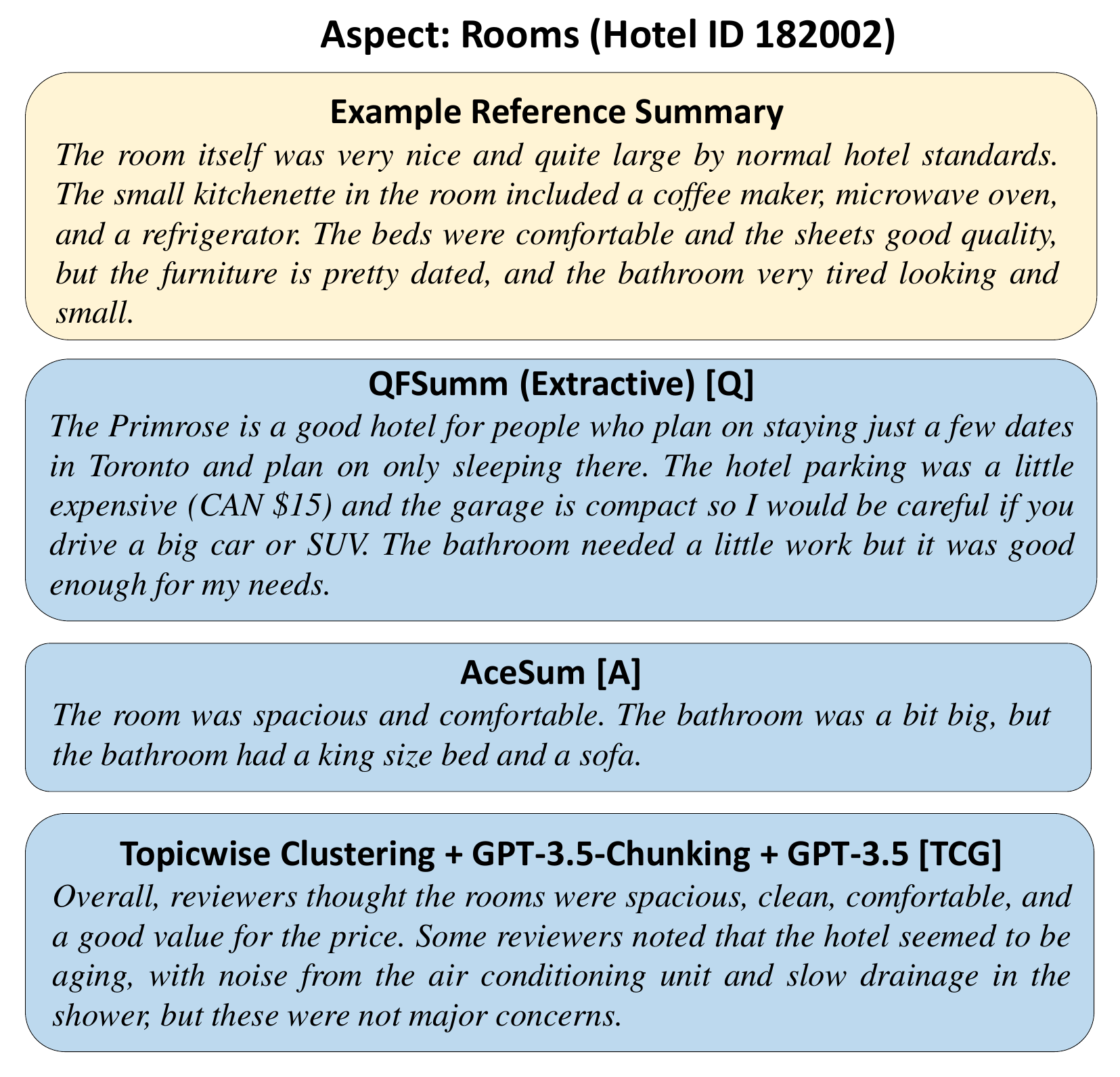}
\caption{Example summaries from TCG, Q, and A, and a reference summary from the SPACE dataset.}
\label{figure:EXS}
\end{figure}

%
\paragraph{Review Stratification (R)} This involves clustering reviews by reviewer scores (given in the dataset) and summarizing each cluster with GPT-3.5.

In addition to extractors, we also utilize \textbf{GPT-3.5-chunking (C)} in some of our pipelines. We segment the sentences from the prior step into non-overlapping chunks, then summarize each individually with GPT-3.5. The results are then concatenated for the next step. 

Our \textit{summarizers} summarize the text one final time to produce the output summary. All of our pipelines use GPT-3.5 as the summarizer. However, we also compare to QFSumm \cite{QFSumm}, AceSum \cite{SPACE} and the model released with FewSum \cite{FewSum}, also referred to as FewSum. 

These building blocks are composed to build various summarization pipelines, which we list in Table~\ref{table:PP}. 
An illustration of one pipeline (TCG) is shown in Figure~\ref{fig:TCGI}. Since topic-wise clustering is unnecessary for FewSum (due to lack of aspects), we only compare G (vanilla GPT-3.5 used to summarize the set of product reviews, truncated to fit if necessary), CG (Chunking + GPT-3.5), QG (QFSumm-long + GPT-3.5), Q (QFSumm), and FS (FewSum) for this dataset.
The table also lists some approaches that are the first stages of pipelines that begin with GPT-3.5-chunking, which we also compare against in Section~\ref{sec:newtools}.

\begin{table}[t]
\includegraphics[width=0.85\linewidth]{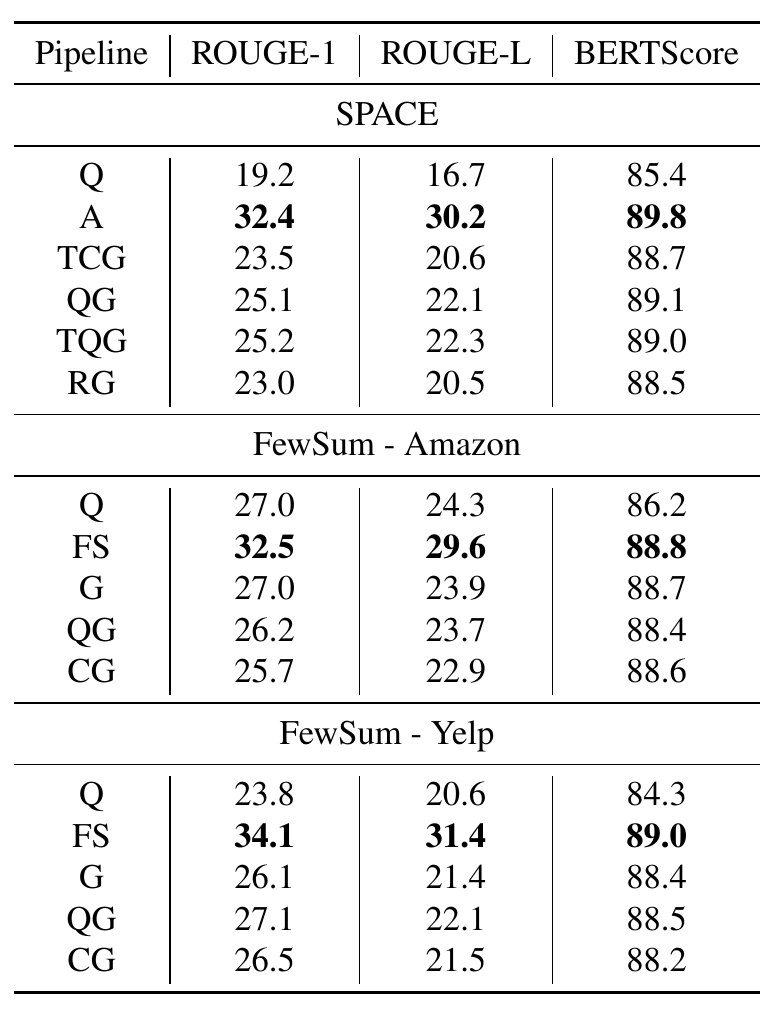}
\caption{ROUGE-1, ROUGE-L, and BERTScore (F1) for the compared models.} 
\label{table:R}
\end{table}
\section{Evaluation}
\begin{table}[H]
\small
  \centering
    \begin{tabular}{c c c } 
    \toprule
      & SPACE & FewSum\\ \midrule
     Average \#reviews per entity & 100.00 & 22.41\\
     Average \#sentences per review & 9.16 & 3.37\\
     Average \#words per sentence & 17.56 & 12.12\\ \bottomrule
    \end{tabular}
  \caption{SPACE and FewSum dataset statistics.}
  \label{table:BINFO}
\end{table}
\subsection{Datasets}

\paragraph{SPACE} \cite{SPACE} involves the summarization of reviews of hotels along the aspects \{\emph{general, rooms, building, cleanliness, location, service, food}\} and provides three human-written summaries for each \emph{(hotel, aspect)} pair. Table~\ref{table:BINFO} shows that the reviews of SPACE are too long to summarize with a non-pipelined system given \texttt{text-davinci-002}'s context window size.
We exclude the \emph{general} aspect from our experiments.

\paragraph{FewSum} \cite{FewSum} contains product reviews from Amazon and Yelp. As opposed to SPACE, FewSum is not aspect-oriented, and the reviews are typically much shorter. 
For many of the products, the combined length of the reviews falls below 900 words, enabling direct summarization with GPT-3.5. 
FewSum provides three gold summaries for only a small portion of the products. Across these two splits, FewSum provides golden summaries for 32 and 70 products in the Amazon and Yelp categories respectively.

We list SPACE and FewSum statistics in Table~\ref{table:BINFO}. 

\subsection{Automatic Eval: ROUGE and BERTScore}

We compute ROUGE \cite{ROUGE} and BERTScore \cite{BERTScore} and show results in Table~\ref{table:R}.

\begin{figure}[t]
\includegraphics[width=0.89\linewidth,trim=7mm 65mm 175mm 30mm]{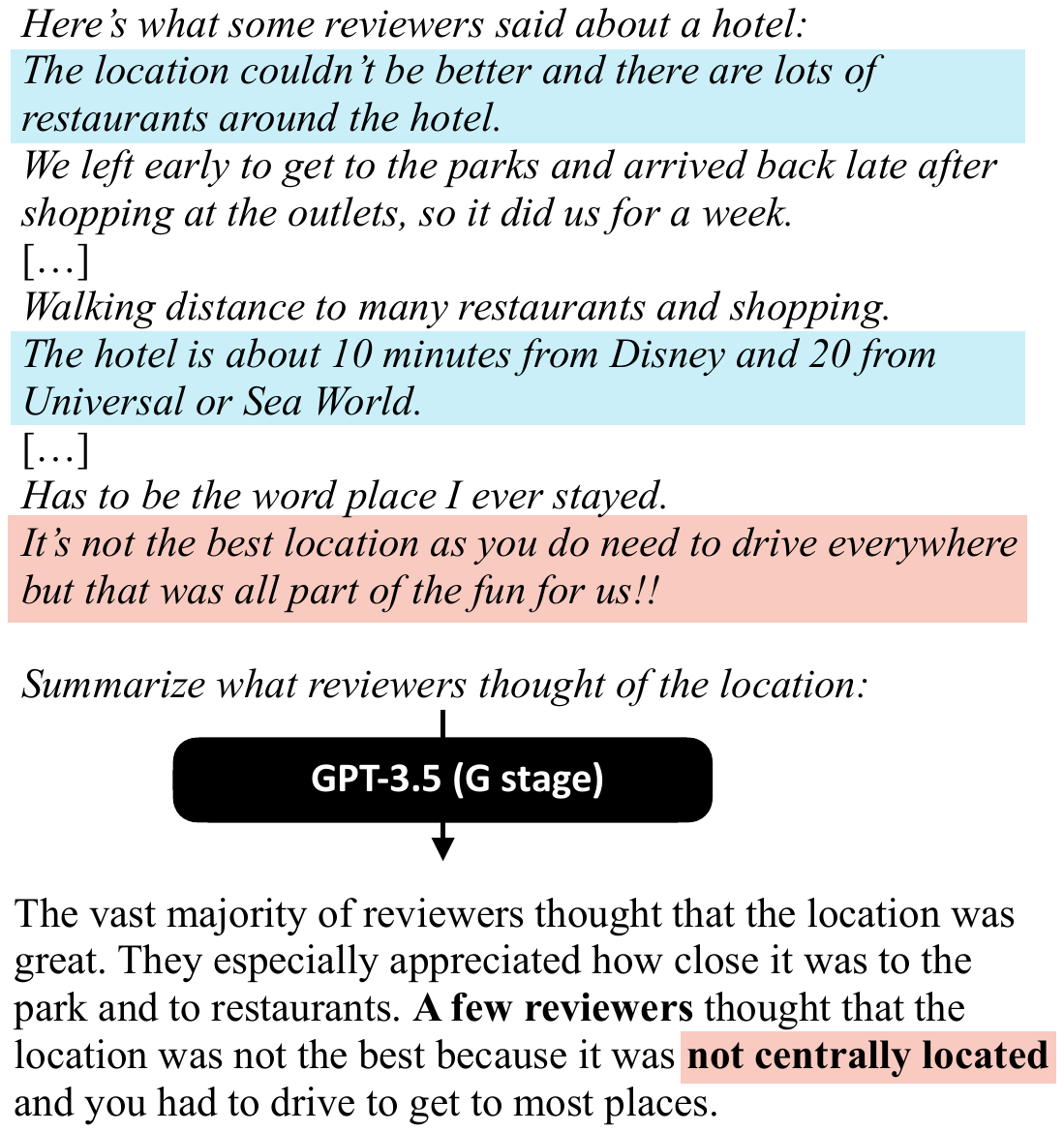}
\caption{Example of errors made by GPT-3.5. The viewpoint of a single reviewer is wrongly expressed as that of a ``few reviewers'' and generalized to the hotel not being centrally located, contradicting other reviews (blue).}
\label{figure:GPTE}
\end{figure}


The BERTScores for AceSum, as well as all GPT-3-related models, are in the range of $88-90$, and differences in performance are unclear. AceSum achieves the highest ROUGE-1 as well as ROUGE-L scores by far, and is followed by TQG and QG. 
QFSumm does particularly poorly on the ROUGE scores.
The scores are all in the same ballpark on FewSum apart from FS, with it being difficult to draw any conclusions. The latter achieves the highest ROUGE-L as well as BERTScore. The GPT-3.5 systems perform slightly better than QFSumm on the Yelp split which we attribute to the smaller combined review lengths of Yelp.

We argue that these scores are not informative and that they are at times unreliable when comparing the quality of two summaries.  ROUGE and BERTScore have been critiqued in prior work as inaccurate indicators of summary quality \cite{SummEval, RPROB1, RPROB2}, particularly as the fluency and coherence of the outputs increase to near-human levels \cite{goyal-gpt3}. Figure~\ref{figure:EXS} demonstrates this by with an example. $n$-gram methods penalize GPT-3.5 for generating summaries in a slightly different style: ``\emph{The reviewers found the rooms to be clean}'' instead of ``\emph{The rooms were clean}.'' Similarly, the extractive nature of QFSumm drives it to produce sentences like ``\emph{We were served warm cookies on arrival}.'' While its selections are factual, they are not completely representative of the review opinions themselves.\ab{How's this?} The actual mistakes in our systems include over-generalization and misrepresentation of viewpoints of popularities thereof, which are not well-represented by matching $n$-grams. Figure~\ref{figure:GPTE} shows an example of such errors. We conclude that metrics benchmarking the summaries on different dimensions are necessary. 

\begin{table}[t]
\centering
\small
\begin{tabular}{ c  | c @{\hspace{0.2cm}} c @{\hspace{0.2cm}} c @{\hspace{0.2cm}} c} 
\toprule
 Pipeline & \makecell{Factuality} & \makecell{Represent-\\ativeness} & \makecell{Faithful-\\ness} & \makecell{Relev-\\ance}\\ \midrule
 TCG & 2.85 & 2.99 & 4.86 & \textbf{4.60}\\
 TQG & 2.86 & 2.95 & 4.83 & 4.32\\
 QG & 2.88 & 2.97 & 4.79 & 3.93\\
 A & \textbf{3.00} & 2.96 & \textbf{4.91} & 3.62\\
 Q & \textbf{3.00} & \textbf{3.00} & 4.88 & 2.30\\
 \midrule
  Maximum & 3 & 3 & 5 & 5\\
 Fleiss-Kappa & \cellcolor{green!40}0.64 & \cellcolor{green!20}0.49 & \cellcolor{green!20}0.49 & \cellcolor{green!40}0.64\\ \bottomrule
\end{tabular}
\caption{Results of Human Evaluation on the SPACE dataset. Colors indicate moderate (light green) and substantial (darker green) agreement, respectively.}
\label{table:HEV}
\end{table}
\begin{table}[t]
\centering
\small
\begin{tabular}{ c  | c @{\hspace{0.2cm}} c @{\hspace{0.2cm}} c @{\hspace{0.2cm}} c} 
\toprule
Pipeline & \makecell{Factuality} & \makecell{Represent-\\ativeness} & \makecell{Faithful-\\ness} & \makecell{Relev-\\ance}\\ \midrule
 G & 2.63 & 2.89 & 4.68 & \textbf{4.98}\\
 CG & 2.72 & 2.95 & \textbf{4.73} & \textbf{4.98}\\
 QG & 2.68 & 2.90 & 4.63 & \textbf{4.98}\\
 Q & \textbf{2.96} & \textbf{2.98} & 4.52 & 4.92\\
 FS & 2.74 & 2.32 & 4.30 & 4.90\\
\midrule
 Maximum & 3 & 3 & 5 & 5\\
 Fleiss-Kappa & \cellcolor{yellow!40}0.26 & \cellcolor{green!20}0.53 & \cellcolor{red!20}0.19 & \cellcolor{red!20}0.15\\ 
\bottomrule
\end{tabular}
\caption{Results of Human Evaluation on the FewSum dataset. Colors indicate moderate (light green), fair (yellow) and slight (red) agreement respectively.}
\label{table:HEVFS}
\end{table}

\subsection{Human Evaluation}
\label{sub:HE}

For a more reliable view of performance, we manually evaluated the summaries of the pipelines TCG, TQG, AceSum (A) and QFSumm (Q) for 50 randomly chosen \emph{(hotel, aspect)} pairs from the SPACE dataset, and G, CG, QG, Q and FS for 50 randomly chosen products (25 each from the \emph{Amazon} and \emph{Yelp} splits) from the FewSum dataset. The axes of evaluation were the attributes established in Subsection~\ref{sub:DD}, namely \emph{Factuality}, \emph{Faithfulness} and \emph{Relevance}. In addition, as we often observed our systems produce summaries of the form ``\emph{While most reviewers thought ..., some said ...}'' to highlight contrasting opinions, we also evaluate on \emph{Representativeness}. Representativeness is a more restricted form of Faithfulness that measures if the more popular opinion was exhibited between two opposing ones. For instance, if four people found the rooms of a hotel clean but two did not, the summary is expected to convey that the former was the more popular opinion.

The three authors of this paper independently rated the summaries along the above axes on Likert scales of 1-3 for both variations of factuality, and 1-5 for faithfulness and relevance. The average scores, along with the Krippendorff's Alpha and Fleiss Kappa scores (measuring consensus among the raters) are presented in Table~\ref{table:HEV}.
Among the compared pipelines, TCG improves upon TQG and QG substantially in terms of relevance. All three have a very high score under Factuality, showing that GPT-3.5 models seldom make blatantly wrong statements. Viewpoints selected by QFSumm are generally faithful, and factual due to their extractive nature, but may include irrelevant statements. 

We list the corresponding metrics for FewSum in Table~\ref{table:HEVFS}. CG tends to perform well, but the consensus is low for Faithfulness and Relevance. FS performs poorly across the board due to hallucinated statements harming its Factuality and bad viewpoint selection resulting in low Faithfulness. The lack of aspects may contribute to the low agreement on FewSum; dimensions such as Relevance may be considered underconstrained, and thus more difficult to agree upon in this setting \cite{kryscinski-etal-2019-neural}.

We remark that all of our systems are achieving close to the maximum scores; the small differences belie that the pipelines all demonstrate very strong performance across the board.

\section{New Tools for Evaluation and Analysis} 
\label{sec:newtools}
Enabling fast automatic evaluation of systems will be crucial for the development of future opinion summarizers. Furthermore, when a large number of reviews are presented to a system, it may be nearly impossible even for a dedicated evaluator to sift through all of them to evaluate a summary. We investigate the question of how we can automate this evaluation using existing tools.

One of the areas where automatic evaluation may help is \textbf{faithfulness}. Since faithfulness represents the degree to which a system is accurate in representing general consensus, it requires measuring the proportion of reviews supporting each claim of a summary. A viewpoint with larger support is more popular and, consequently, more faithful. Our key idea is to use entailment as a proxy for support. Past work \cite{goyal-durrett-2021-annotating,SUMMAC} has used Natural Language Inference (NLI) models to assess summary factuality by computing entailment scores between pairs of sentences.

\begin{figure}[t]
    \centering
    \includegraphics[width=0.95\linewidth,trim=3mm 90mm 95mm 20mm]{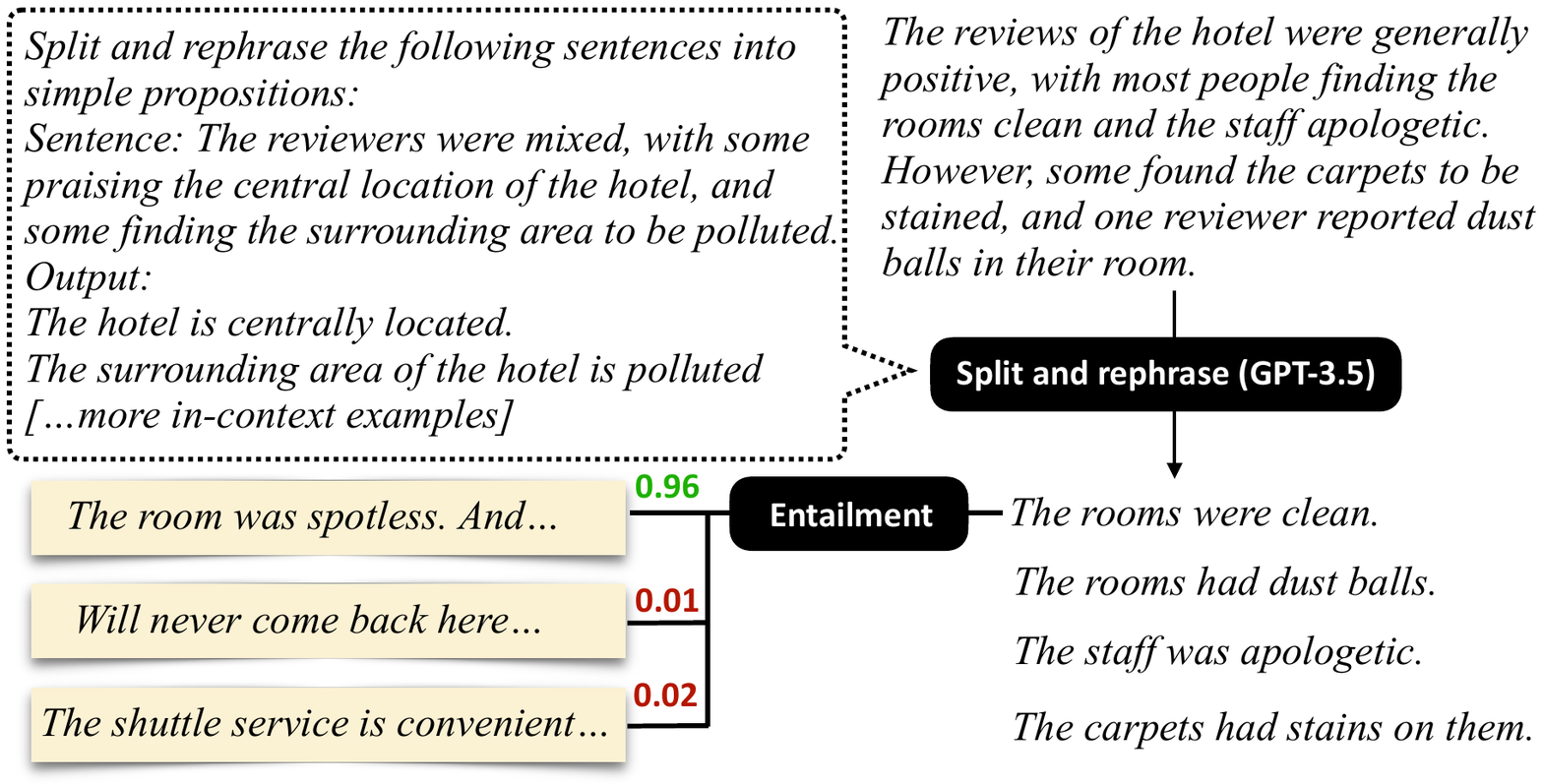}
    \caption{Per-sentence entailment scores are calculated by taking the maximum among the various candidates.}
    \label{fig:ent}
\end{figure}

However, the summaries produced by GPT-3.5 and related pipelines often consist of compound sentences that contrast two viewpoints. In addition, GPT-3.5 prefers to say ``\emph{The reviewers said...}'' instead of directly stating a particular viewpoint. We found these artifacts to impact the entailment model. We use a split-and-rephrase step to split these sentences into atomic value judgments by prompting GPT-3.5 as shown in Figure~\ref{fig:ent}. We then use the zero-shot entailment model from SummaC \cite{SUMMAC} to compute the entailment scores for these atomic value judgments. 
Similar to the approach in the SummaC paper, we observe that a summary statement is factual when strongly entailed by at least one sentence and thus select the top entailment score of each summary sentence as its \textbf{factuality score}, and aggregate this score to produce per-system numbers. 
The choice of the model as well as that of using GPT-3.5 for the split-and-rephrase step are explained further in Appendix \ref{ap:SR}, and the relevant metric of abstractiveness is discussed in Appendix~\ref{ap:abs}.

A system could potentially game this metric by producing relatively ``safe'' statements (like \emph{most reviewers found the rooms clean}). We therefore also want to evaluate \textbf{genericity}.

\subsection{Terminology}

The set of sentences in the summary of the reviews of a hotel $h \in  \mathcal{H}$ w.r.t aspect $a \in \mathcal{A}$ is called $S_{h,a}$. Passing these to the split-and-rephrase step gives us a set of split sentences $Z_{h,a}$. For any two sentences $s_1, s_2$ we denote the entailment score of $s_2$ with respect to $s_1$ according to the SummaC-ZS \cite{SUMMAC} model by $e(s_1,s_2) \in [-1.0, 1.0]$. A score of $1.0$ indicates perfect entailment while that of $-1.0$ denotes complete contradiction. Finally, we denote by $N_n(s)$ the (multi-)set of $n$-grams (with multiplicity) of the sentence $s$. In particular, $N_1(s)$ is the set of words in the sentence $s$.

\subsection{Evaluation of Entailment}
We first evaluate whether entailment is effective at identifying the support of the mentioned viewpoints by human evaluation. The three authors of this paper marked 100 random pairs (50 each from SPACE and FewSum) of sentences and assertions entailed with a score above $0.5$ on the scale of $0-2$. Here, $2$ indicates that the assertion is completely supported, and $1$ that the assertion's general hypothesis is supported, but some specifics are left out. The average score of the selection across the raters was \textbf{1.88} with a Fleiss Kappa consensus score of 0.56 (moderate agreement). Many of the lower-rated entailed sentences also had lower entailment scores (closer to 0.5). The score illustrates that the precision of the entailment approach is high.

\begin{table}[t]
\centering
\includegraphics[width=0.95\linewidth]{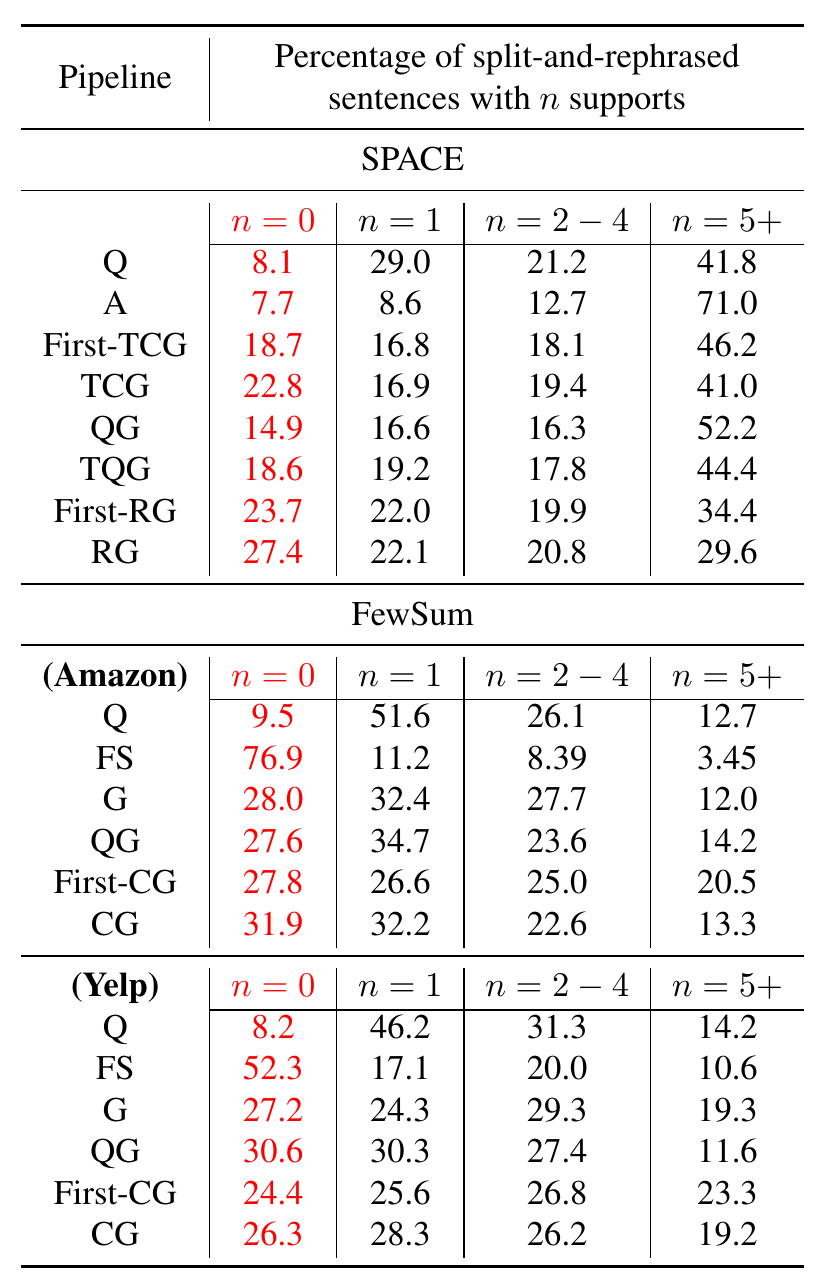}
\caption{Percentages of split-and-rephrased sentences binned according to support sizes, for all compared pipelines. The threshold used is $\tau = 0.75$.}
\label{table:P}
\end{table}

\subsection{Faithfulness: Support Set Sizes}
\label{sub:FF}

We propose an entailment metric for determining how the viewpoints in the summary reflect the consensus of the input. 
%
%
We first compute per-sentence entailment scores as shown in Figure~\ref{fig:ent}. 
For each sentence of the split-and-rephrased summary, we measure the number of review sentences that entail it with a score greater than a threshold $\tau = 0.75$ (the ``support'' of the sentence).
This threshold was determined based on manual inspection.
We bin these counts into $0,1,2-4$ and $5+$.  
The frequencies of the bins are converted to percentages and listed in Table~\ref{table:P}.
FS performs poorly due to presenting hallucinated viewpoints, and repeated summarization slightly hurts CG on the Amazon split. G and CG outperform other methods on the Yelp split, likely because it has fewer reviews per product than Amazon, making it much likelier for the combined reviews of a product to fit in a manageable number of words. The ``pure'' GPT-3.5 systems generally perform well on the short review sets of FewSum. 
As we move to the long combined lengths of the reviews on SPACE, however, the pure GPT-3.5 pipelines fall behind in terms of faithfulness. 
Repeated summarization causes a major dip from First-TCG to TCG, indicating that this is not effective for long-form inputs. 
QG outperforms other GPT-3-related pipelines by a large margin. 
%
\begin{table}[t]
    \centering
    \small
      \begin{tabular}{cc|ccc}
        \toprule
        Pipeline & \makecell{Average\\Top Score} & Pipeline & \multicolumn{2}{c}{Average Top Score}\\
        \midrule
        \multicolumn{2}{c|}{SPACE} & \multicolumn{3}{c}{FewSum}\\
        \midrule
        Q & 91.59 &   & \textbf{(Amazon)} & \textbf{(Yelp)}\\
        A & \textbf{92.49} & Q & \textbf{85.29} & \textbf{86.62} \\ 
        First-TCG & 84.96 & FS & 24.36 & 47.23 \\  
        TCG & 82.06 & G & 65.81 & 68.59 \\ 
        QG & 87.50 & QG & 67.63 & 65.04 \\
        TQG & 84.68 & First-CG & 68.34 & 69.86 \\
        First-RG & 81.54 & CG & 66.43 & 68.58 \\
        RG & 79.85 &  &  & \\ 
        \bottomrule
      \end{tabular}
    \caption{The average Top Score for each pipeline on the SPACE and FewSum datasets.}
    \label{table:top-score}
\end{table}
As we saw in human evaluation, however, QG may include some irrelevant viewpoints in this process. 
Abating this behavior by performing a topic-clustering step first brings its numbers down to a level comparable with First-TCG, which is still more faithful than the TCG pipeline.  
AceSum has the largest number of statements with $5+$ supports on the SPACE; however, as we will see later, many of its summaries are very generic, and support for them can be easily found among the large number of reviews. Q has the smallest percentage of statements with no support because it is extractive.

\subsection{Factuality: Top Score}
\label{ap:TS}
As depicted in Figure~\ref{fig:ent}, averaging the per-sentence entailment scores (first per-summary, then per-system) gives us the \emph{Top Score} metric. The average top score is a proxy for factuality since true statements will typically be strongly entailed by at least one sentence of the reviews.
%
We list the computed average top scores in Table~\ref{table:top-score}.
FS performs poorly on FewSum in terms of Factuality. The numbers for other systems are similar, with QG and CG performing best on the Amazon and Yelp splits.
However, on the longer inputs of SPACE, the differences in factuality become more apparent. 
In particular, to reconcile similar but distinct viewpoints, repeated summarization leads to a type of generalizing that hurts the factuality of TCG and TG.
Among the GPT-3.5 pipelines, QG performs the best, followed by TQG. 
TQG yet again delivers performance comparable to First-TCG and therefore presents a reasonable trade-off with some gains on factuality and increased relevance. 
%

\begin{table}[t]
\centering
\includegraphics[width=\linewidth]{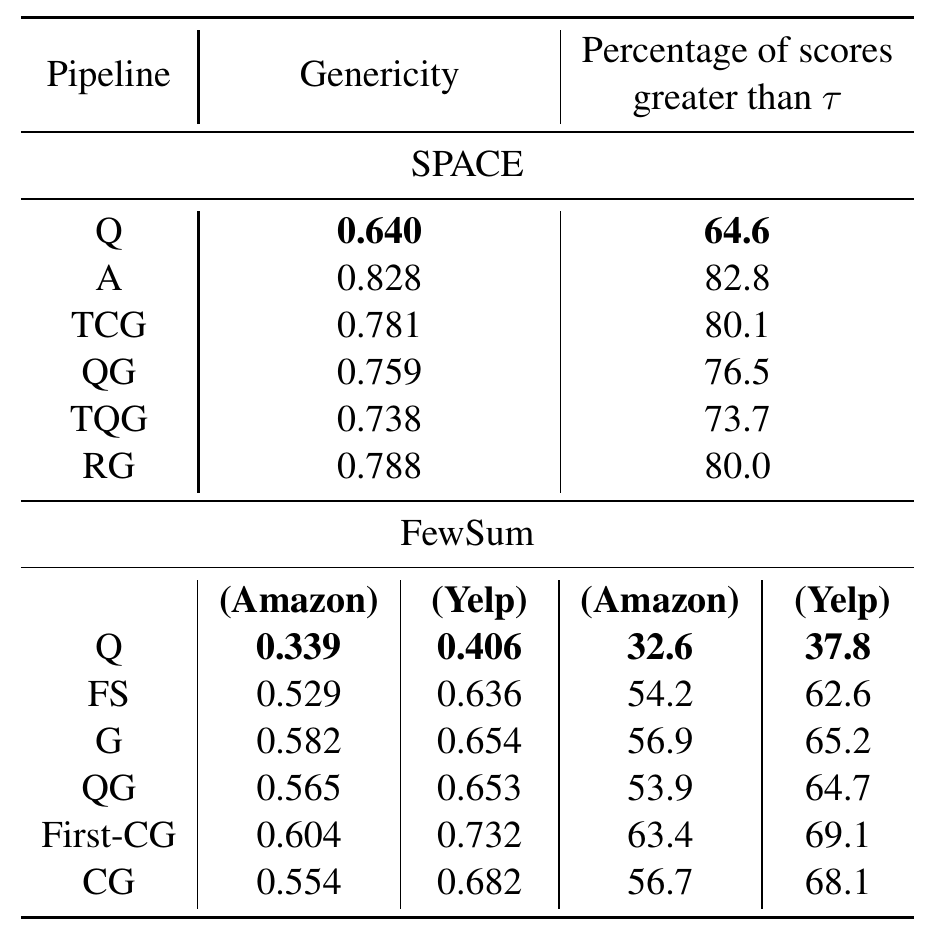}
\caption{Semantic genericity based on entailment, along with the raw percentage of scores above the threshold. The threshold used is $\tau = 0.5$.}
\label{table:genericity}
\end{table}

\subsection{Genericity}
\begin{table}[t]
\centering
\small
\centering
\begin{tabular}{c c | c c  c} 
    \toprule
    Pipeline & \makecell{Average\\IDF} & Pipeline & \multicolumn{2}{c}{Average IDF}\\
    \midrule
    \multicolumn{2}{c|}{SPACE} & \multicolumn{3}{c}{FewSum}\\
    \midrule
    Q & \textbf{12.00} & & \textbf{(Amazon)} & \textbf{(Yelp)}\\
    A & 5.77 & Q & \textbf{4.38} & \textbf{4.33}\\
    TCG & 8.40 & FS & 3.16 & 3.26\\
    QG & 6.93 & G & 3.02 & 2.93\\
    TQG & 7.82 & QG & 3.10 & 2.93\\
    RG & 8.87 & CG & 3.00 & 2.86\\
    \bottomrule
\end{tabular}
\caption{Measurement of lexical genericity. Average IDF (larger is better) for the compared pipelines. The FewSum pipelines report lower ranges for average IDF due to fewer total number of documents.}
\label{table:lex-genericity}
\end{table}

As mentioned before, we want to measure whether reviews contain largely generic statements like \emph{the service was helpful}, which are likely to be faithful and factual but not very useful to a user of a system. 

We first focus on \emph{semantic} genericity, i.e. the use of statements generally applicable to other products/services in the same class. On the other hand, \emph{lexical} genericity involves the overuse of generic words and is tackled next.
Our approach to measuring semantic genericity employs the observation that generic sentences from a summary are often widely applicable and thus likely to be strongly entailed by statements from other summaries. We calculate the similarity \texttt{sim}$(S,S')$ of two sets of sentences using the averaged top score, as Figure~\ref{fig:ent} shows. Similarly, we also measure the fraction \texttt{frac}$(S, S', \tau)$ of sentences whose top score exceeds a threshold $\tau$. Equation \ref{eq:G2a} computes the average similarity score between sentences that belong to two reviews by the same system but different (\emph{hotel, aspect}) pairs (normalizing by the number of pairs $N$). Equation \ref{eq:G2b} computes the corresponding metric based on \texttt{frac}.
\begin{equation}
   \begin{aligned}
    G  & = \frac{1}{N}\sum\limits_{(h,a) \neq (h', a')} \texttt{sim}(Z_{h,a}, Z_{h',a'})
    \label{eq:G2a}
    \end{aligned}
\end{equation}
\begin{equation}
   \begin{aligned}
    F_\tau & = \frac{1}{N}\sum\limits_{(h,a) \neq (h', a')} \texttt{frac}(Z_{h,a}, Z_{h',a'}, \tau)
    \label{eq:G2b}
   \end{aligned}
\end{equation}
 We report these two metrics in Table~\ref{table:genericity}. On the short inputs of FewSum, all GPT-3.5 pipelines give similar results, with FewSum being slightly less generic. Moving to SPACE, however, the range of scores becomes much wider. Forced to reconcile disparate opinions during repeated summarization, TCG and RG produce generic summaries, although AceSum is the most generic. We note that pre-extraction with QFSumm and Topic-wise clustering help QG and TQG remain less generic. 

To measure \emph{lexical genericity}, we use the sentences from \textit{all} summaries on the corresponding dataset as the set of documents to calculate an averaged Inverse Document Frequency (IDF) of the summaries, with stopwords removed and stemming applied. Since generic words are likely to occur more frequently and therefore have a low IDF, a smaller score indicates higher genericity. The scores calculated this way are listed in Table~\ref{table:lex-genericity}. As expected, QFSumm is highly specific due to being extractive. We observe that AceSum generates summaries that over-use generic words, in line with our prior observations. We also note that pre-extraction with QFSumm helps with lexical genericity as it did with semantic genericity. Finally, on FewSum, we observe that FS does better than every other pipeline apart from Q. This bolsters our previous claim that its low Factuality and Faithfulness scores were due to hallucinated, but specific, viewpoints.
\subsection{Correlation with Human Judgments}
\begin{table}[t!]
    \small
    \centering
    \begin{tabular}{c|c|c}
        \toprule
        Evaluation Axis & \makecell{Entailment-Based\\Metric} & \makecell{ROUGE}\\
        \midrule
        Factuality & \cellcolor{green!20}0.36 & \cellcolor{red!20}0.05\\
        Faithfulness & \cellcolor{green!20}0.29 & \cellcolor{red!20}-0.03\\
        \bottomrule
    \end{tabular}
    \caption{Spearman Correlation Coefficients of our metrics and ROUGE with human judgments.}
    \label{tab:SPRCC}
\end{table}

\noindent
Our entailment-based approaches set out to measure Factuality and Faithfulness; how well do these correlate with our human evaluation? We compute Spearman's rank correlation coefficient on the human-annotated SPACE examples with the averaged annotator scores, as the consensus among rater scores was high on that dataset. 
In particular, we use the average of the Factuality scores among the raters as the net human score on Factuality on an example and the mean score on Faithfulness as that for Faithfulness. Correspondingly, we consider the Top Score metric as the automatic measurement of Factuality and the percentage of statements with $3$ or more supports as Faithfulness. We list the obtained Spearman correlation coefficients in Table~\ref{tab:SPRCC}. 
While there is room for stronger metrics, the fact that the introduced metrics correlate with human judgments better than ROUGE provides an encouraging signal that these target the factors of interest.

\section{Related work} 
\paragraph{Text Summarization} Historically, most work tackling text summarization has been \emph{extractive} in nature \cite{Ext-old-1, Ext-old-2, Ext-old-3, Ext-old-4}, with more recent work applying pre-trained extractive systems to this task \cite{Ext4, Ext3, Ext2, Ext1, QFSumm}. 
\textit{Abstractive} approaches \cite{Ext-old-3, Abs-old-1, Abs-old-3} to summarizing reviews have become more successful in recent years \cite{Abs4,Abs2,Abs3,Abs1}. We follow in this vein, capitalizing on the strength of GPT-3.5.

\paragraph{Multi-Stage Summarization} Most systems of both types are now end-to-end \cite{SS1,SS2,QFSumm}. However, multi-stage approaches \cite{mohit2018,MS2, MS1} like ours have recently shown great promise. 
For instance, \citet{MS2} extracts relevant evidence spans and then summarizes them to tackle long documents. 
Recursive summarization has been explored in \cite{wu-books} for book summarization, but involved fine-tuning GPT-3.5 to the task. Other approaches such as the mixture-of-experts re-ranking model \citet{EX} can be considered as a two-step approach where the \texttt{combine} function ranks and filters the outputs of the first stage. 

\paragraph{Evaluation Metrics} The domain of news summarization has recently seen interest in using factuality/faithfulness for evaluation \cite{QuestEval, FactCC, tang2023understanding}. In news, faithfulness and factuality are quite similar, as news articles usually do not present incorrect information or conflicting opinions. Opinion summarization is therefore quite distinct in this regard, and a separate treatment of factuality and faithfulness is sensible. For the same reason, although unified approaches to evaluating text generation \cite{unified1, unified2} are useful, more targeted metrics are likely to be more informative for opinion summarization specifically.

\paragraph{Aspect-Oriented Summarization} In addition to opinion summarization \cite{SPACE}, aspect-oriented summarization has also been explored in other domains of NLP \cite{aspectbased1,aspectbased2}. However, as highlighted above, opinion summarization differs from news summarization with respect to desired characteristics, and this work focuses specifically on those issues.

\section{Conclusion}

In this work, we show that GPT-3.5-based opinion summarization produces highly fluent and coherent reviews, but is not perfectly faithful to input reviews and over-generalizes certain viewpoints. 
ROUGE is unable to capture these factors accurately. We propose using entailment as a proxy for support and develop metrics that measure the faithfulness, factuality, and genericity of the produced summaries. 
Using these metrics, we explore the impact of two approaches on controlling the size of the input via pre-summarization on two opinion summarization datasets. With the reasonably sized inputs of FewSum, GPT-3.5 and CG produce faithful and non-generic outputs. However, as we move to long-form review summarization, the factuality and faithfulness of these approaches drop. A pre-extraction step using QFSumm helps in this setting but leads to generally shorter and more generic summaries; a topic clustering step can then make summaries less generic and more relevant at a small cost to faithfulness and factuality.
We hope that our efforts inspire future improvements to systems and metrics for opinion summary evaluation. 

\section*{Limitations}

Our study here focused on the most capable GPT-3.5 model, \texttt{text-davinci-002}, at the time the experiments were conducted. We believe that models like ChatGPT and GPT-4, as well as those in the future, are likely to perform at least as well as these, and if they improve further, the metrics we have developed here will be useful in benchmarking that progress. However, significant further paradigm shifts could change the distribution of errors in such a way that certain of our factors (e.g., genericity) become less critical. In addition, the latest iterations of GPT have a much greater input window size, which help them digest much larger swaths of text in one go and potentially make our pipelined approaches less needed in certain settings.

Furthermore, the \texttt{text-davinci-002} model is fine-tuned with data produced by human demonstrations. The precise data used is not publicly available, so it is difficult to use our results to make claims about what data or fine-tuning regimen leads to what failure modes in these models.

Recent work has noted that language models may be susceptible to learning biases from training data \cite{lm-bias-1, lm-bias-2, lm-bias-3}, and this phenomenon has also been observed for GPT-3.5 \cite{gpt3-bias}.
We did not stress test the models studied for biases and furthermore only experimented on English-language data. 
\par 
When properly used, the summarization models described in this paper can be time-saving.
However, as noted above, summary outputs may be factually inconsistent with the input documents or not fully representative of the input, and in such a case could contribute to misinformation. 
This issue is present among all current abstractive models and is an area of active research.

\section*{Acknowledgments}

This work was partially supported by NSF CAREER Award IIS-2145280, a grant from Open Philanthropy, a gift from Salesforce, Inc., and a gift from Adobe. Thanks as well to the anonymous reviewers for their helpful comments.

\bibliography{refs.bib}
\bibliographystyle{acl_natbib}

\pagebreak
\appendix
\begin{figure*}[t]
\includegraphics[width=0.95\linewidth]{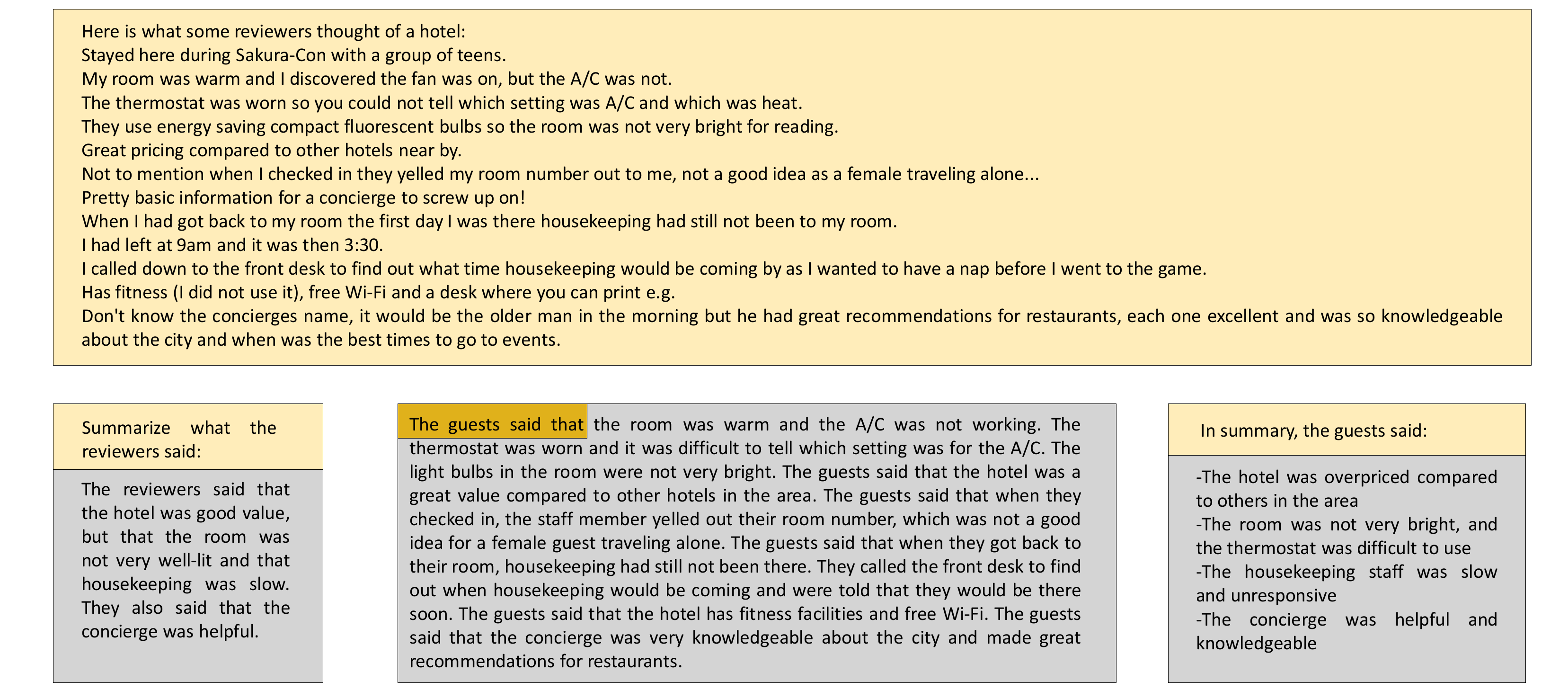}
\caption{Aspects of summarization such as verbosity or the format of output are affected by the specific wording of the prompt. We use the leftmost prompt, ``\emph{Summarize what the reviewers said.}''}
\label{figure:PE}
\end{figure*}
\begin{figure*}[t]
\includegraphics[width=0.95\linewidth]{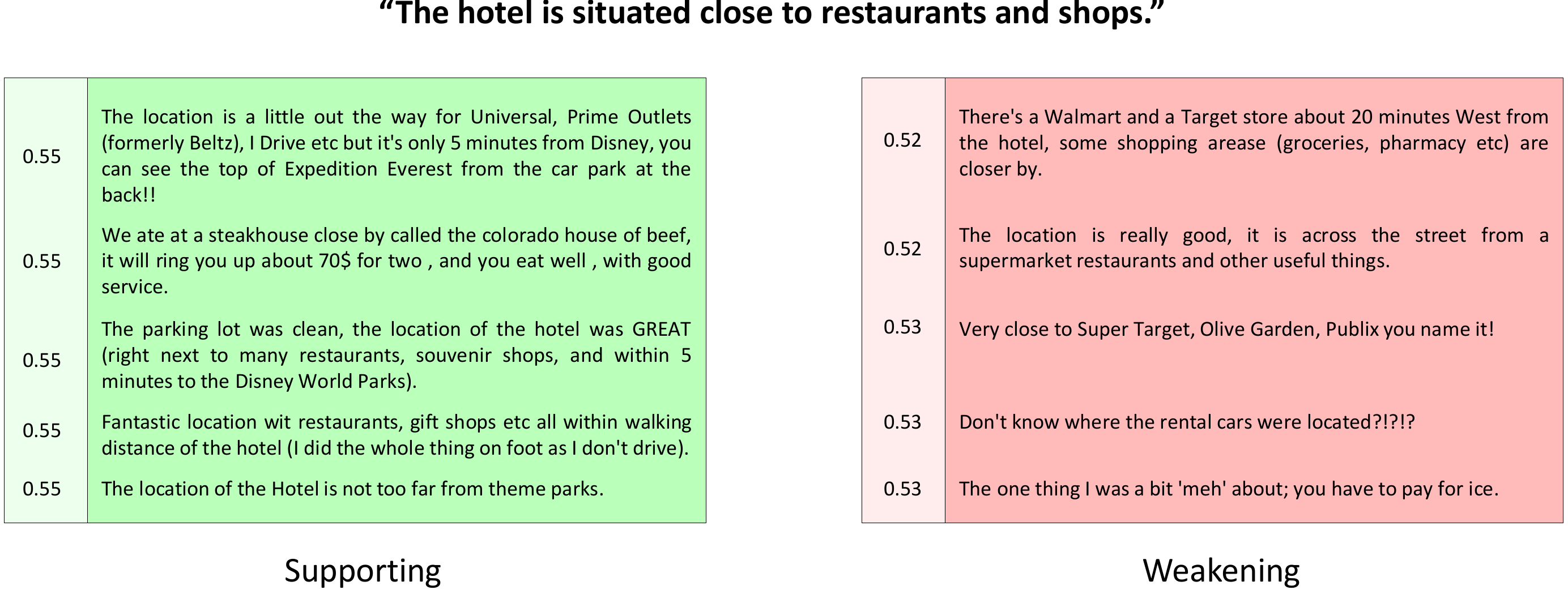}
\caption{The top 5 supporting and weakening sentences from the reviews for the statement ``\emph{The hotel is situated close to restaurants and shops}'' as found by the \textbf{Conv} SummaC model. The corresponding entailment scores are included in parentheses. We see that the scores are very close to each other and that the ``weakening'' statements do not weaken the statement at all. These issues led us to use the zero-shot model instead.}
\label{figure:SC}
\end{figure*}
\begin{figure*}[t]
\includegraphics[width=0.95\linewidth]{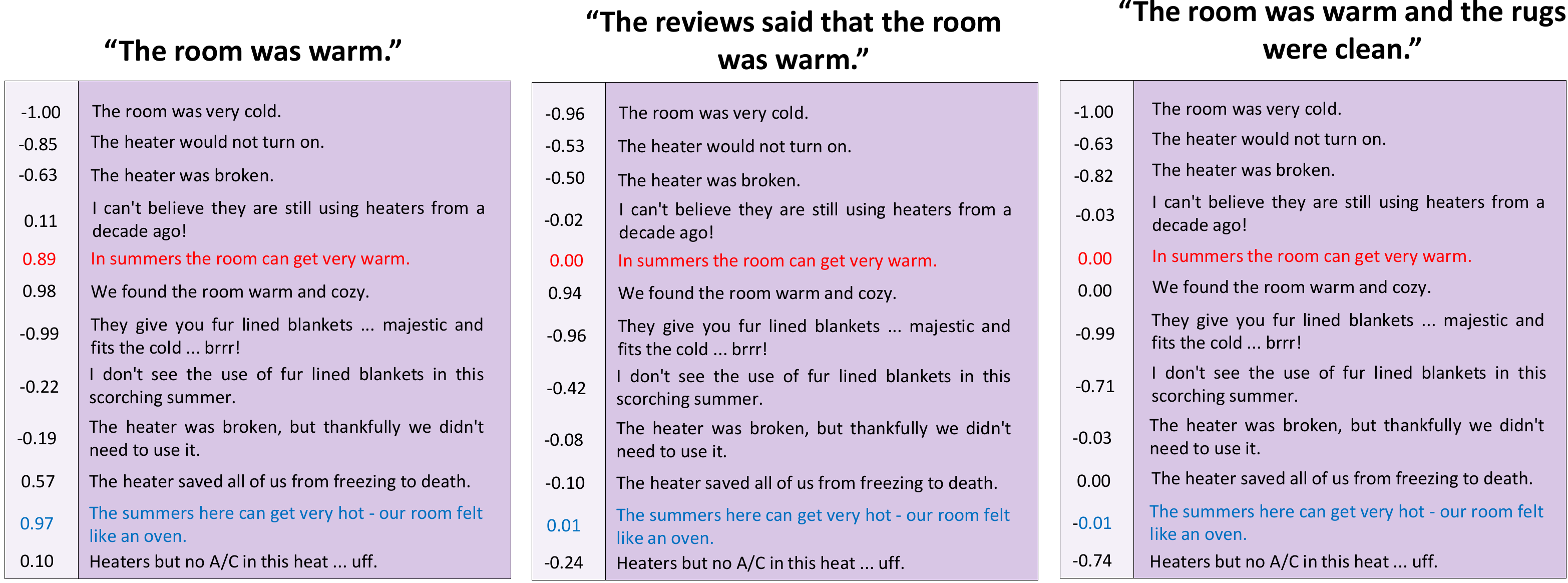}
\caption{The scores of three statements with respect to a set of sentences, highlighting the issues with directly using the model output to compute entailment scores. Scores rounded to three decimal places are included before the corresponding sentences, with important lines highlighted in color. We note that \textcolor{red}{quoting} a proposition as said by someone else or having \textcolor{blue}{multiple propositions} in the same sentence serve to cloud entailment scores.}
\label{figure:SRR}
\end{figure*}
\section{Pipeline Details}
\label{ap:PD}
\subsection{Details of the Infrastructure, Models, and Datasets Used}

\paragraph{Computational Resources} All experiments were run on a machine equipped with an Intel Xeon W-2123, and utilized a TITAN RTX GPU with a 24 GB memory. We estimate the total computational GPU budget to be roughly 100 GPU-hours.

\paragraph{Model Sizes} QFSumm \cite{QFSumm} is a fine-tuned version of BERT and therefore has 110M parameters. The FewSum model from \cite{FewSum} has 25.1M parameters including the plug-in network. AceSum \cite{SPACE} has a combined total of 142M parameters between the Controller Induction Model and Opinion Summarization Model. We use the VitC variant of the entailment model SummaC-ZS \cite{SUMMAC}, which relies on the ALBERT-xlarge architecture with 60M parameters. For all models, we used the default parameters as reported in \citet{QFSumm}, \citet{FewSum}, \citet{SPACE}, and \citet{SUMMAC}. Consequently, no hyperparameter search was necessary. All models have been publicly released under the MIT License on GitHub by the respective authors.

\paragraph{Datasets and Evaluation} Both the SPACE and FewSum datasets consist of reviews in English. The former consists of reviews of hotels, and the latter product reviews from Amazon and service reviews from Yelp. We are using pre-existing datasets that are standard in opinion summarization. Through our human evaluation, we did not see any personal identifying information or offensive content in the reviews we assessed. 
All of our human evaluation experiments were performed once by the authors, and we report the Krippendorff's Alpha and Fleiss Kappa scores as measurements of consensus.
We used ROUGE with the default settings.\footnote{The \texttt{rouge.properties} file at \url{https://github.com/kavgan/ROUGE-2.0}} We used NLTK's \cite{NLTK} WordNet \cite{wordnet} lemmatizer as the lemmatizer where needed. Sentence splitting was done using the \texttt{sent\_tokenize()} function of NLTK.
\subsection{Details of the Configurations and Prompts}
Here we provide more details of the configuration and/or prompts used for various models. Below, GPT-3.5 refers to the \texttt{text-davinci-002} model.
\paragraph{QFSumm and QFSumm-long (Q)} QFSumm allows one to specify the number $n$ of sentences to extract from the reference text to shape into a summary. We use $n=3$ (the default setting) for QFSumm (summarizer) and $n=35$ for QFSumm-long (extractor). On the SPACE dataset, we use the aspect-specific keywords from \citet{QFSumm} to pass to the model. On the FewSum dataset, however, the set of relevant keywords may be drastically different across examples. Therefore, for each product, we pass $5$ randomly chosen reviews to GPT-3.5 with the prompt consisting of the reviews and the directive ``\emph{Output up to eight comma-separated keywords that capture these reviews most saliently:}''. The produced keywords are then used with QFSumm to summarize the reviews.
\paragraph{GPT-3.5 Topic Clustering (T)} The prompt we use is ``\emph{Describe the topic of each sentence in one word}'', followed by three examples and then the sentence whose topic is to be determined. We then map the produced words to their corresponding normalized GloVe \cite{GLOVE} vectors, which are then mapped to the closest aspects in terms of L2 distance. This is functionally equivalent to using cosine similarity as the vectors are normalized.
\paragraph{GPT-3.5 Chunking (C)} We strive for the length of the chunks (in sentences) to be both as close to each other and to 30 as possible; thus, when there are $l$ sentences total to be chunked, we take $c = \lceil\frac{l}{30}\rceil$ to be the number of chunks, and allocate $\lfloor \frac{l}{c} \rfloor$ sentences to each chunk (except the last one, which may have fewer).
\paragraph{Review Stratification (R)} If a cluster's length exceeds GPT-3.5's upper limit at this stage, it is truncated to the maximum number of sentences that fit. 
\paragraph{GPT-3.5 (G)} When used as a summarizer, we feed the penultimate set of sentences to GPT-3.5 with the prompt \textit{``Summarize what the X said of the Y:,''} where X is either ``\emph{reviewers}'' or ``\emph{accounts}'' based on whether GPT-3.5-chunking was used so far. Y is the aspect being summarized (SPACE) or just ``\emph{Product}'' (FewSum).  The preamble is either \textit{``Here's what some reviewers said about a hotel:''} or \textit{``Here are some accounts of what some reviewers said about the hotel''} in the case of SPACE. The word ``\emph{hotel}'' is replaced by ``\emph{product}'' for FewSum.

\section{Entailment and Decomposition}
\label{ap:SR}
In line with our motivation, we would like to be able to use an NLI (Natural Language Inference) model to retrieve entailment scores of the produced summaries with respect to the input reviews. We tested several approaches including BERTScore, due to it being trained on entailment/contradiction pairs, but finally settled on using the zero-shot model from SummaC \cite{SUMMAC} to produce the entailment scores. SummaC is already becoming a standard evaluation tool for summarization factuality. We chose to forego the trained ``Conv'' SummaC model as we found that it did not generalize well to the kind of data we were working with. Specifically, two common issues were that (1) the range of scores assigned to the sentences from the reviews was very small, and (2) sometimes (especially for the most weakening statements) the scores assigned to the sentences seemed arbitrary and did not make a lot of sense. In comparison, the zero-shot model had neither of these issues. This issue is highlighted in Figure~\ref{figure:SC}.

Further, a proposition X is typically not judged by models to entail statements of the form ``\emph{The reviewers said X}'', or ``\emph{X and Y}'', where \emph{Y} is another proposition. Accordingly, the entailment scores are not very high for these two cases. We highlight this in Figure~\ref{figure:SRR}.
Thus, we decide to split and rephrase all sentences of the produced summary to simple value propositions for all entailment-related metrics. Note that here rephrasing also includes removing any attribution such as ``\emph{The guests said...}''. We considered several models to this end, including BiSECT \cite{BiSECT} and ABCD \cite{ABCD}, but found two common issues with all of them:
\begin{itemize}
    \item The split sentences maintained the words from the original sentences, so a sentence such as ``\emph{The food was received well but it was served late}'' would have one output part as ``\emph{It was served late}'', which requires a round of entity disambiguation to follow the split-and-rephrase step.
    \item These models do not remove attribution of viewpoints as we would like.
    \item A statement such as ``\emph{I liked the setting of the movie but not its cast}'' produces one of the outputs as ``\emph{Not its cast}'', which does not make any sense by itself. 
\end{itemize}
Thus, we utilize GPT-3.5 to perform the split-and-rephrase task, with few shot prompting used to illustrate the removal of attribution and other desired characteristics. We also experimented with having separate steps for split-and-rephrase and found no significant difference in the outputs or quality thereof. We utilize the split-and-rephrased sentences for all of the automatic metrics that involve entailment of any sort.
\begin{table}[t]
\centering
\includegraphics[width=\linewidth]{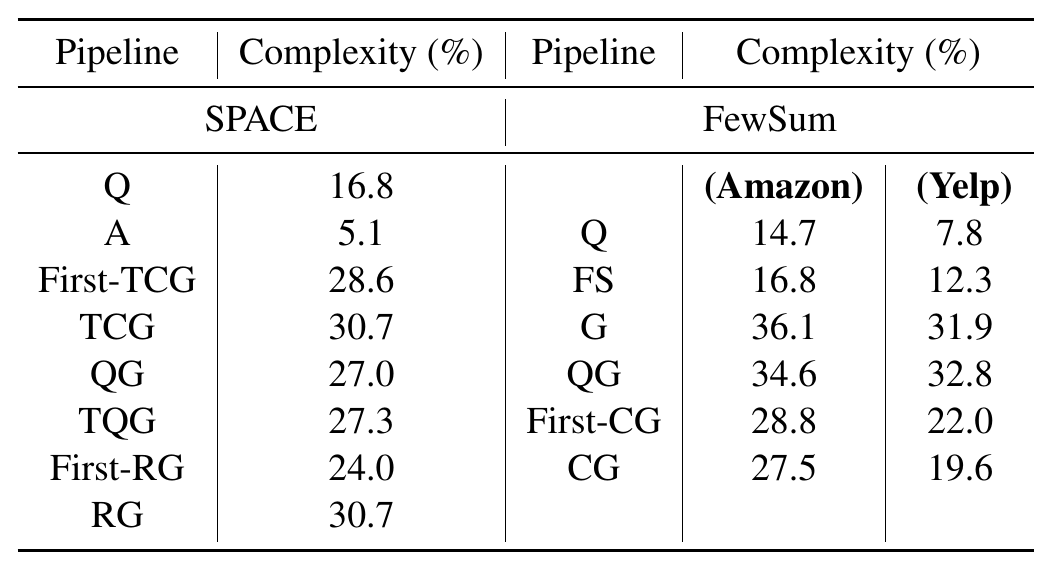}
\caption{Complexity as measured by the percentage of contrasting sentences.}
\label{table:C}
\end{table}
\begin{figure*}[t]
\centering
\includegraphics[width=0.95\linewidth]{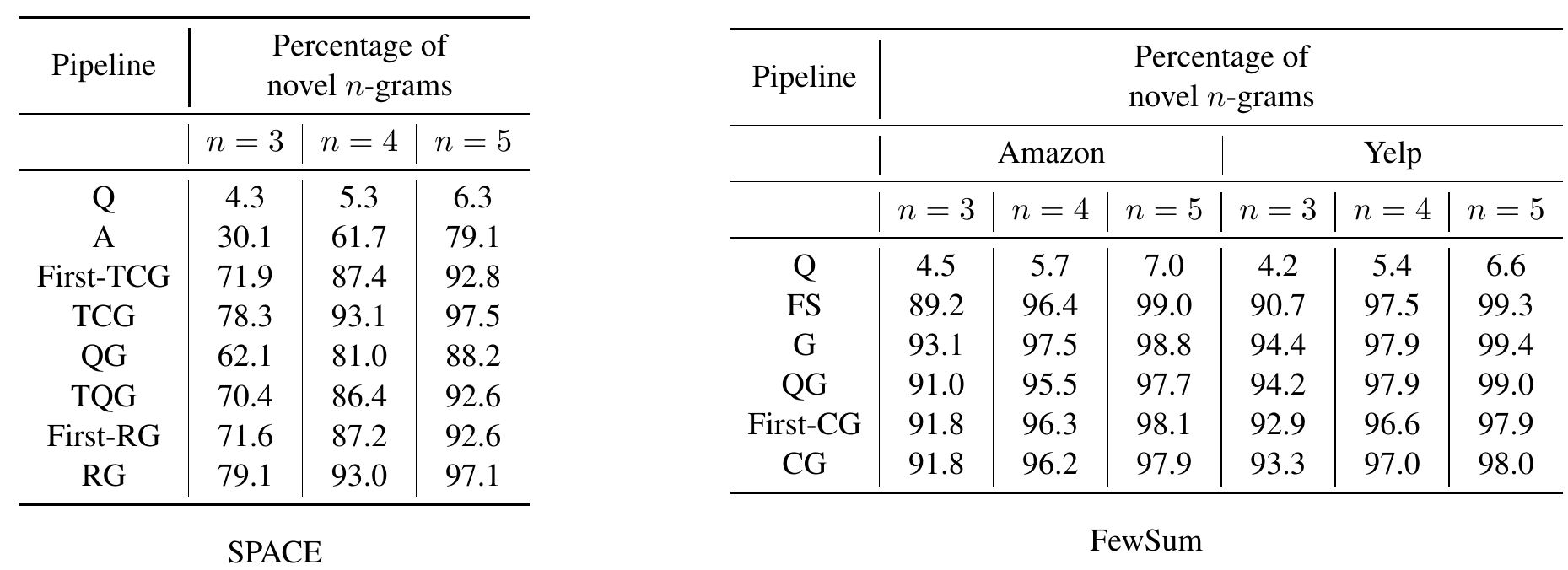}
\caption{Abstractiveness as measured by the percentage of novel $n$-grams when compared with the source reviews} 
\label{table:A}
\end{figure*}
\section{Measuring Complexity}
One of the challenges of opinion summarization is that sentences may contrast opinions: ``\emph{Most reviewers liked the service, but there were a few complaints about sluggish response times.}'' We quantify the percentage of simple and contrasting statements in the model outputs since it is subtly related to the extent of expression of opposing viewpoints. We use the original (non-split) sentences for this purpose and classify a sentence as contrasting if it contains one or more words from the set $\mathcal{K} = $ \texttt{\{'while', 'but', 'though', 'although', 'other', 'others', 'however'\}}, as Equation \ref{eq:C} depicts.  We present these percentages in Table~\ref{table:C}.
\begin{equation}
    C = \frac{\sum\limits_{h \in \mathcal{H}, a \in \mathcal{A}}\sum\limits_{s \in S_{h,a}}\mathbbm{1}(N_1(s) \cap \mathcal{K} \neq \emptyset)}{\sum\limits_{h \in \mathcal{H}, a \in \mathcal{A}} |S_{h,a}|}
    \label{eq:C}
\end{equation}
We note that AceSum produces the smallest percentage of contrasting statements. We see that topic-wise clustering pushes up the number of contrasting statements for QG. We hypothesize that this is because when bringing together statements with the same topics in a cluster two opposing statements are likelier to fall into the same chunk. In cases where two opposing statements fall into different chunks, say X and Y, the chunks are likely to each contain statements similar to others in the same chunk. Thus, the summaries of those chunks are likely to be highly contrasting and thus increase the above measure even more for the final stage, as is observed above for TCG.
\begin{figure}[t]
    \centering
    \includegraphics[width=\linewidth]{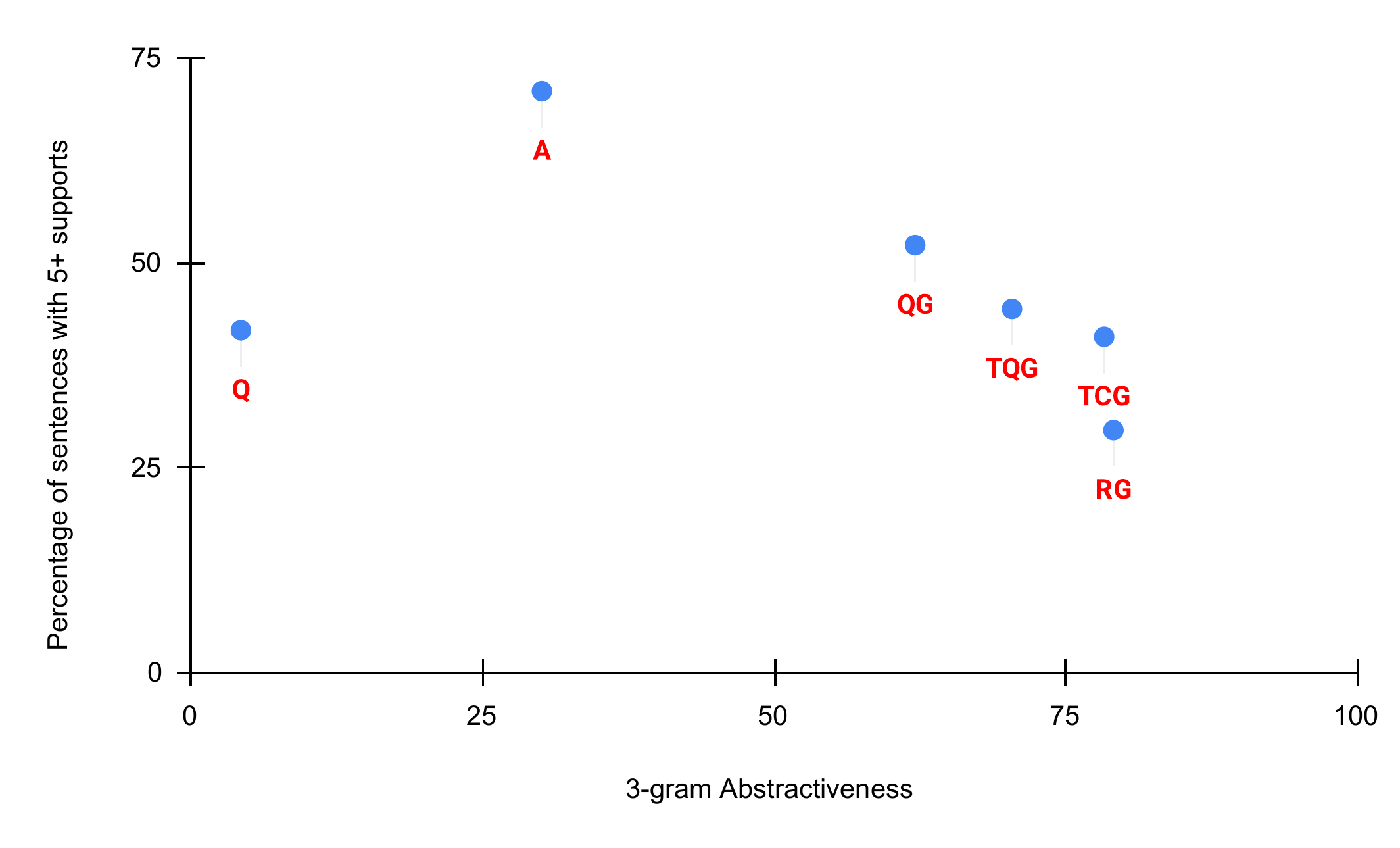}
    \caption{Average Top Score v/s Abstractiveness on the SPACE dataset.}
    \label{fig:TSA}
\end{figure}
\section{Abstractiveness}
\label{ap:abs}
We further investigate how the choice of the pipeline affects abstractiveness. To measure this, we calculate the percentage of $n$-grams in the summaries that do not appear in the input reviews for $n \in \{3,4,5\}$. For this, we use the original (non-split) sentences from the output summaries. The results are tabulated in Table~\ref{table:A}.

Since QFSumm is a purely extractive model, it is no surprise that Q has low abstractiveness. The numbers are non-zero due to some quirks of QFSumm about splitting into sentences - this leads to some partial sentences ending up next to each other. The next stand-out is that A has very low abstractiveness. This is in line with our observation that even though AceSum is abstractive, it tends to highly generic observations such as ``\emph{The rooms were clean}'', which very likely appear almost verbatim in some user reviews. We also observe that QG has a relatively low abstractiveness and that topic clustering drives up abstractiveness. We suspect that the above is a result of GPT-3.5 simply mashing together some sentences when presented with chunks containing highly disparate sentences (since it is hard to find a common thread among them), which promotes extraction over abstraction. Another observation is that multi-GPT-3.5 pipelines (TCG and RG) are more abstractive than single-GPT-3.5 ones since there are two rounds of abstraction as opposed to one. All the GPT-3.5-derived pipelines are highly abstractive in the case of FewSum, and slightly more so than FS. This is unsurprising since the combined length of the reviews in the case of FewSum is much smaller when compared to SPACE, and therefore there are relatively fewer propositions to compress into general statements. Motivated by \citet{Ladhak}, we display the line graph of the average Top Score vs.~$3$-gram Abstractiveness for the SPACE dataset in Figure~\ref{fig:TSA}. The trio of QG, TQG, and TCG define the best frontier on the Factuality-Abstractiveness tradeoff, followed by RG, then A and Q.
\end{document}